\documentclass[conference]{IEEEtran}
\IEEEoverridecommandlockouts
\usepackage{cite}
\usepackage{amsmath,amssymb,amsfonts}
\usepackage{algorithmic}
\usepackage{graphicx}
\usepackage{textcomp}
\usepackage{xcolor}
\usepackage{url} 
\DeclareUnicodeCharacter{00A0}{ } 
\DeclareUnicodeCharacter{200B}{ } 
\def\BibTeX{{\rm B\kern-.05em{\sc i\kern-.025em b}\kern-.08em
    T\kern-.1667em\lower.7ex\hbox{E}\kern-.125emX}}
\begin{document}

\title{Intelligent Electric Power Steering: Artificial Intelligence Integration Enhances Vehicle Safety and Performance
\\
}

\author{\IEEEauthorblockN{1\textsuperscript{st} Vikas Vyas}
\IEEEauthorblockA{\textit{Autonomous Driving Department} \\
\textit{Mercedes-Benz Research and Development North America}\\
Sunnyvale, CA, USA \\
talkwithvikas@gmail.com}
\and
\IEEEauthorblockN{2\textsuperscript{nd} Sneha Sudhir Shetiya}
\IEEEauthorblockA{\textit{IEEE Senior Member} \\
Troy, MI, USA \\
sneha.shetiya@ieee.org}

}

\maketitle

\begin{abstract}
Electric Power Steering (EPS) systems utilize electric motors to aid users in steering their vehicles, which provide additional precise control and reduced energy consumption compared to traditional hydraulic systems. EPS technology provides safety,control and efficiency.. This paper explains the integration of Artificial Intelligence (AI) into Electric Power Steering (EPS) systems, focusing on its role in enhancing the safety, and adaptability across diverse driving conditions. We explore significant development in AI-driven EPS, including predictive control algorithms, adaptive torque management systems, and data-driven diagnostics. The paper presents case studies of AI applications in EPS, such as Lane centering control (LCC), Automated Parking Systems, and Autonomous Vehicle Steering, while considering the challenges, limitations, and future prospects of this technology. This article discusses current developments in AI-driven EPS, emphasizing on the benefits of improved safety, adaptive control, and predictive maintenance. Challenges in integrating AI in EPS systems. This paper addresses cybersecurity risks, ethical concerns, and technical limitations,, along with next steps for research and implementation in autonomous, and connected vehicles.
\end{abstract}


\section{Introduction}
Electric Power Steering has transitioned from traditional hydraulic systems to advanced  electronically controlled mechanisms. EPS systems apply an electric motor to assist the driver in steering the vehicle, replacing hydraulic pumps and hoses. 

The core components of a modern EPS system typically include:
Electric motor
Torque sensor
Steering Angle Sensor
Electronic Control Unit (ECU)
Reduction gear

The ECU analyzes inputs from ADAS sensors to process the appropriate level of assistance, which is then applied by the electric motor through the reduction gear.

Artificial Intelligence, with capacity to handle  large amounts of data and make instantaneous decisions, is effective to enhance EPS systems. AI comprises various technologies such as Neural Networks, Deep Learning, Machine Learning, Fuzzy Logic, and Evolutionary Algorithms.In terms of EPS,  AI enables systems to adapt to lane centering, learn driver preferences, and anticipate steering needs before they emerge.

The complex data processing and pattern recognition in AI will transform EPS technology features, enabling real-time adaptation to road conditions, vehicle speed, and driver behavior. EPS is the backbone of advanced driver assistance systems (ADAS) and autonomous vehicles (AVs). It is interesting to see how AI technologies are creating new opportunities for EPS, which is generating interest among automotive manufacturers and researchers..

In the current automotive domain, there is a high demand for electric, autonomous vehicles with energy efficiency. EPS is a crucial control unit in modern steering systems. These systems provide steering force by processing filtered data from sensors and user inputs, supported by steering angle and torque sensors. EPS enhances safety, adjusts to driving conditions in real-time, and improves overall vehicle performance through the integration of AI. Where this technology will lead next is an exciting path.

The role of Artificial Intelligence (AI) in Electric Power Steering (EPS) is transforming the automotive field by improving vehicle safety, security, efficiency, and the overall driving experience. At the same time, by using machine learning algorithms and real-time data analytics, AI enhances performance, personalizes the driving experience, predicts potential failures, and ensures a safe driving environment.

\section{Review of Existing Research}
This research synthesizes findings on the current state of AI-driven EPS, focusing on contributions to vehicle safety, stability, adaptive performance, and predictive maintenance. It also discusses current challenges, such as cybersecurity, data analysis, and ethical considerations in autonomous applications.
\subsection{AI in Electric Power Steering Systems}
    The Role of AI in EPS systems is a rising  area of study, with researchers focusing on real-time control and improving the system’s flexibility in dynamic driving environments. Abeyratne, Hoang, and Ng (2021) mentioned that AI's integration into EPS systems is primarily through deep learning (DL) and machine learning (ML)  algorithms, which enable the analysis of data from various sensors to predict optimal steering responses based on driver inputs,environmental factors and road conditions. Such algorithms enable EPS to provide more intuitive steering feedback and reduce driver effort \cite{B1}.
    
    Additionally, Chen and He (2022) focus on  how AI in EPS allows for an adaptive and flexible response to different driving situations, improving driver comfort while ensuring a stable and secure steering experience. Their work highlights the potential of  reinforcement learning, which allows EPS to learn from driver behavior, adjusting system responses to enhance control without compromising safety. These studies underscore the potential of AI-enhanced EPS systems to offer customizable, context-sensitive driving experiences  \cite{B2}.  

\subsection{Enhancing Vehicle Safety through AI-Driven EPS}
    The research reveals the critical role of AI-driven EPS systems in safety-critical applications, including lane-keeping assistance and collision avoidance. Gupta and Luo (2020) found that AI algorithms can enable EPS systems to predict and adjust to potential hazards by analyzing steering angle, vehicle speed, road condition, and obstacle proximity. This allows the steering system to support collision avoidance maneuvers and provide lane-keeping maneuvers\cite{B3}.
    
   Kim and Park (2023) highlight the potential of AI-enhanced EPS to  improve vehicle stability and performance improvement, particularly in challenging conditions such as rain or snow. Their research reveals  that EPS systems using AI algorithms adapt angle, torque and steering input based on real-time data, preventing oversteer or understeer. These safety benefits are critical in emergency situations, where rapid and precise adjustments are needed to maintain control\cite{B4}.

\subsection{AI-driven Predictive Maintenance and Diagnostics}
\begin{itemize}
    \item Predictive maintenance has become an important area of focus, where EPS systems equipped with AI can foresee potential component failures and recommend timely maintenance. Chen and He (2022) explain that machine learning models can analyze sensor data to detect abnormal patterns, such as deviations in torque or irregularities in motor performance. This approach enables proactive servicing, which minimizes the likelihood of sudden failures and extends component lifespans \cite{B2}.
    Wang and Wu (2020) discuss the potential for AI-driven diagnostics to reduce repair costs and vehicle downtime by accurately diagnosing issues before they escalate. This predictive capability is crucial in commercial fleet management, where timely maintenance reduces operational disruptions. The integration of AI in EPS not only enhances safety but also adds economic value by reducing maintenance costs and extending the lifespan of the steering system \cite{B9}.
\end{itemize}

\subsection{Technical and notable Challenges in AI-Driven EPS Systems}
\begin{itemize}
    \item Although AI in EPS brings substantial benefits, the research highlights notable challenges. Martin and Wong (2022) discuss cybersecurity as a major concern, as the connectivity of AI-driven EPS systems increases vulnerability to hacking. Since EPS is directly involved in vehicle control, securing these systems against unauthorized access is crucial\cite{B6}.
    Another critical area of concern is the ethical dilemma surrounding autonomous decision-making. Rahman and Singh (2023) note that AI-driven EPS systems may face scenarios where they must make decisions in unavoidable accidents. The ethical
    considerations of these decisions are complex, as they involve balancing risks to the driver, passengers, and other road users. The research  highlights the requirements for regulatory supervision and ethical frameworks to guide the development of such systems, assuring safety while maintaining responsibility\cite{B7}.
    Furthermore, the power and computational demands of AI models remain a challenge. Zhang and Kumar (2021) indicate that edge computing is one potential solution, as it enables on-board data processing, reducing the energy demands and latency associated with cloud-based processing. Nevertheless the AI* -driven EPS proactively identify and Address potential issues, leading to improving vehicle stability, vehicle performance and reducing maintenance cost\cite{B10}.
\end{itemize}

\subsection{Future Directions and Opportunities}
\begin{itemize}
    \item Overall this research highlights the potential of AI in EPS systems, particularly in enhancing safety and performance. Future direction should be development of low-power AI algorithms to mitigate the energy demands of data processing in AI-driven EPS systems. To address privacy concerns and improve model performance, Collaborative learning enables vehicles to collaborate and learn from each other without sharing sensitive data. Robust cybersecurity measures are essential as vehicles become increasingly connected and autonomous, providing protection against cyber attacks, in edge computing, cybersecurity measures, and ethical frameworks. AI-driven EPS can enable a broader range of applications, providing a new pathway for the next generation of intelligent, autonomous, and connected vehicles\cite{B11}.
\end{itemize}

\section{AI Integration in EPS: Safety and Performance enhancement}
\begin{itemize}
      \item In EPS systems, the traditional mechanical parts are replaced or supported by electric motors and sensors, which provide precise steering control. This basic structure provides an ideal platform for AI integration, where sensors collect data on torque,steering angle, vehicle speed, position, and driver input. This is real time data analysis from an AI algorithm to help to optimize the steering input, make real time adjustments, ensure safe and efficient vehicle operation. Machine learning techniques, such as reinforcement learning and deep learning enable the system to adapt to environment conditions, driving style and road conditions. The integration of AI into the EPS system brings significant advantages such as real time decision making , predictive maintenance and robust redundancy mechanisms. 
      The incorporation of AI into EPS systems brings several significant advantages, including real-time decision-making, predictive maintenance, and redundancy mechanisms
\end{itemize}
\begin{itemize}
 \item Steering Angle and Torque Sensors- Enhancing Accuracy and Redundancy :  Steering angle and torque sensors are essential components of an EPS system, which provides the critical data on the driver's input and the road conditions. AI integration enhances these input values by continuously monitoring the sensor data, detecting inconsistencies, and providing real-time corrections.The redundancy of AI-powered sensors enables a fail-operational approach, allowing the system to operate safely by relying on backup data from other sensors if one fails. This increases the accuracy of steering adjustments and ensures the system's robustness against sensor failures.
 \item Real-Time Data Processing :  AI-driven EPS systems can analyze vast amounts of data from steering angle, torque sensors, and other vehicle sensors like cameras, radar, and lidar to enable precise and immediate steering adjustments. These systems can detect important changes in driver inputs, vehicle dynamics, and road conditions,ensuring optimal performance and safety across various driving scenarios.
  \item Adaptive Steering - Different driving Environments:  AI enables EPS systems to adapt to various personalized driving environments. For example, in urban settings, steering responsiveness can be adjusted for smoother handling at low speeds. On high-speed highways, AI can modify steering assistance to make hard maneuvers easier and reduce driver fatigue. These adjustments are essential for enhancing both the personalized driving experience and driver safety.
  \item Enhanced Safety - Redundancy Mechanisms : Although redundancy mechanisms are already available in current EPS systems, the integration of AI makes crucial contributions to fail-operational capabilities. By filtering data from multiple sensors, including steering angle and torque sensors, AI smoothly transitions to other systems to maintain control and avoid fail-safe scenarios if one sensor fails or produces erroneous data. This fail-operational approach is particularly critical in autonomous and semi-autonomous vehicles, where system reliability is essential for safe operation.
 \item Predictive Maintenance : AI also plays a vital role in predictive maintenance by continuously monitoring the health and performance of key components such as the electric motor, steering angle sensors, and torque sensors. AI-driven EPS systems enhance vehicle reliability and reduce downtime for repairs, thereby improving overall efficiency by identifying potential issues before they lead to system failure.
\end{itemize}
\subsection{The Role of AI in EPS}
\begin{itemize}
    \item Predictive Steering Algorithms : AI algorithms in EPS systems can predict the need for steering adjustments based on real-time data from various sensors. These predictive algorithms analyze different factors such as driver behavior, lateral acceleration, steering torque, vehicle speed, road conditions, and yaw rate. By processing this data through neural networks, the system can anticipate required steering inputs milliseconds before they are needed.
    The predictive capabilities of these systems often rely on Long Short-Term Memory (LSTM) networks or Recurrent Neural Networks (RNNs).
    A typical LSTM cell used in predictive steering might be described by the following equations\cite{B14} :
    \begin{align*}
    ft =\sigma(W_f \cdot [h_{t-1}, x_t] + b_f) \qquad (1)\\
    it =\sigma(W_i \cdot [h_{t-1}, x_t] + b_i) \qquad (2)\\
    \tilde{C}_t = \tanh(W_C \cdot [h_{t-1}, x_t] + b_C) \qquad (3) \\
    C_t = f_t\cdot C_{t-1}+i_t\cdot\tilde{C}_t \qquad (4)\\
    o_t = sigma(W_o\cdot [h_{t-1}, x_t] + b_o) \qquad (5)\\
    h_t = o_t \cdot \tanh(C_t) \qquad (6)
    \end{align*}
   Where $f_t$, $i_t$, and $o_t$ are the forget, input, and output gates respectively, $C_t$ is the cell state, $h_t$ is the hidden state, $W$ and $b$ are weight matrices and bias vectors, $\sigma$ is the sigmoid function, and $*$ denotes element-wise multiplication\cite{B14}.
    \item AI algorithm for Adaptive Torque Control:  Machine learning algorithms for adaptive torque control learn from driver preferences and behaviors to optimize steering effort. These systems often apply reinforcement learning algorithms, such as Deep Deterministic Policy Gradient (DDPG), to optimize steering effort.
    \item AI-Driven Fault Detection and Diagnostics:  AI models analyze sensor data for the predictive maintenance of EPS systems, detecting potential failures before they occur by continuously monitoring system performance and component health. These predictive maintenance systems often use detection algorithms, such as Isolation Forests or One-Class SVMs, to identify unusual patterns in sensor data that may indicate impending failures.
     \item Deep Learning for Steering Control: Deep learning models, such as Convolutional Neural Networks (CNNs) and Recurrent Neural Networks (RNNs), are trained on large datasets collected from real-world driving scenarios. This training allows them to generalize across a wide range of driving conditions and forecast optimal steering actions.
      \item Reinforcement Learning for Dynamic Handling: These reinforcement learning algorithms learn from the feedback of previous actions, optimizing steering performance over time [5]. For instance, a Q-learning algorithm might be used to learn optimal steering adjustments for weather conditions and different road types.
\end{itemize}
 \subsection{Key AI Applications in EPS}
 \begin{itemize}
 \item Lane Keeping Assist (LKA): 
 Lane Keeping Assist systems use AI to predict vehicle drift and adjust steering to keep the vehicle centered in its lane. A typical LKA system might use a convolutional neural network (CNN) to process camera images and detect lane markings\cite{B8}.
\item Automatic Parking Assist :
These AI models in automatic parking systems use real-time data from cameras and sensors to create a 3D map of the parking environment, calculate the optimal steering path, and autonomously maneuver vehicles in tight parking spaces
\item Steering in Autonomous Driving : 
AI algorithms can make complex steering decisions in real-time, adapting to traffic, pedestrians, and road conditions by integrating multiple data streams from GPS, cameras, radar, and LiDAR sensors.
\end{itemize}
 \subsection{Essential Requirements for AI-Driven EPS Systems}
  For data quality and labeling, there are many companies in the Industry which provide data for dedicated scenarios. Data generation is crucial for any Autonomy feature for any sensor including steering ECUs. This data has to reflect all Operational Design Domain conditions including certain corner cases. This vigorous training helps eliminate false positives thus avoiding hard brakes and false steering command values. Ensuring testing and validation as per required standards ensures safety.
 Once data is received, the next step is to label them. Data annotations are tricky and nowadays we have algorithms developed to automate the same. With the evolution of Generative AI, we can use Generative adversarial networks which utilize the concept of Generator and Discriminator to generate more synthetic data for these purposes\cite{B11}.
Particularly for steering and braking ECUs the response time is quick, of the order of milliseconds. Thus, it's crucial to train the model heavily before deployment to have comfort when driving to have faster response times. Fine tuning any model which follows back propagation principles, having dropout layers and ReLU functions activated works wonders \cite{B12}.
Once the data is trained and labeled and the model is tuned for necessary parameters, the next step is to ensure that appropriate faults are getting detected and MRM (Minimum Risk Maneuver) works as expected. MRM is the action that the vehicle performs if a fault is triggered. These faults are categorized into low, fail safe, critical bugs and accordingly directed into the code(Health monitor) which governs the movement of the vehicle.
\subsubsection{Challenges in AI-Driven EPS Systems}
\begin{itemize}
 \item Higher computing requirements for complex data : 
 
Computing power is essential as the software package size increases. Consider the example if we have to flash a 4 GB firmware onto a steering ECU to enable its functionality including diagnostics of  the Ecu, interconnection to interfaces and so on. Flashing this maybe a few minutes. As a next step, we add our AI software to the package and it's now 16 GBs. The same takes almost an hour to flash. This is not reasonable in a product because then a vehicle must be stationary for so long for the update to occur. Having an HPC with partitions enabled to store multiple packages and parallel flashing and compatibility checks ensures speedy process.
Certificates are necessary to ensure secure diagnostics and flashing. Cybersecurity is also important to prevent malware attacks and hacking.
As crucial it is to have an HPC, it's also important to have a redundant chassis. This ensures proper failsafe operations, commissioning and deployment. These are the must haves to have a working AI-driven EPS in the product.
\item  Cybersecurity and Privacy Concerns :
Data privacy in AI powered vehicles is a debatable topic. One has to ensure proper certificates are enabled to ensure security of the user data by avoiding any breaches in the security through hacking and malicious malware attacks.
\item Technical and Ethical Concerns : 
ML models are still in development and have not achieved 100
\item Data Processing and Power Consumption :
Any application using data needs storage capabilities and processing capabilities. All the software data that gets flashed on the ECU needs power, high computational power to be precise. There is a goal in the Automotive Industry to maximize fuel efficiency and this nature of AI curbs it.
\end{itemize}

\subsubsection{Impact on Autonomous vehicles with different levels of Autonomy}

The integration of AI in EPS systems holds profound implications for both autonomous and conventional vehicles.When it comes to Autonomous vehicles, one needs to consider different levels of Autonomy as Autonomous Driver Assistance Features vary from level to level. Level 3 onwards , human intervention drastically reduces and completely is eliminated from Level 5. In level 4 as well certain safety aspects are to be considered but is still heavily autonomous. These steer-by-wire systems and Electric power steering systems play a huge role in the design of Autonomous vehicles. Use Cases vary for autonomous driving based on the usage for Trucks for long haul (Ex: Kodiak, Aurora) to cars( Ex: Waymo, Cruise).
The former need highway driving scenarios and platooning while the latter need to be well equipped to transport passengers in urban as well as highway scenarios. These change the sensors on the vehicle including the responsiveness of the steering system.
In conventional vehicles, AI-driven EPS systems enhance safety and comfort through features like adaptive steering, lane-keeping assistance, cruise control, Anti Lock braking feature and collision avoidance. These advancements significantly improve the driving experience, particularly in challenging driving conditions like snow, blackice and terrains.
\begin{itemize}
\item OEMs providing AI enhanced steering: 
    Tesla: This has the Autopilot system that integrates Machine Learning into its steering systems.
    Toyota: Another big OEM which is focusing on advancing its steering systems to incorporate AI.
    General Motors: GM has two, Cruise and Super Cruise which focus on developing self-driving cars. These require sophisticated EPS technologies to make it to production.
    Ford: Ford is another OEM heavily investing in driver assistance systems to utilize intelligent steering.
   Tier-1 Suppliers:
   To meet above OEM demands, suppliers have also improved their technology both in terms of security and safety to ensure proper functioning of the steering systems in case any fault occurs.
   Bosch: They are the suppliers of steering systems to OEMs and design them as per the Functional Safety standards (ISO26262)
   Continental: This is a major supplier with experience in Automotive fields for different ECUs(Electronic Control Units) and are currently developing solutions to incorporate AI and machine learning.
   Nexteer Automotive: They are the global leaders in steering systems . Specific details about their implementation in production is not known but their AI-powered steering systems enhance safety, comfort and convenience
\end{itemize}

\subsubsection{AI-Driven Electric Power Steering in comparison to traditional drive by wire systems}

There are 4 major areas where drive by wire systems are not preferred and thus AI Driven Electric Power Steering is preferred.
\begin{itemize}
\item Reliability and safety: They lack the tactile feedback offered by EPS which causes users to lose confidence and control on the vehicle.
\item Expensive: Complexity is more here than in EPS systems. Thus more connectivity and hardware make the integration expensive by increasing manufacturing cost.Maintenance and repair incur additional charges too. In case of AI-driven EPS, remote software updates and dependency on software reduces cost.
\item Acceptance and Trust: It's difficult to adapt to new technologies . In EPS, the mechanical connection between steering wheel and wheels is retained. Thus giving everyone a familiar experience.
\item Regulatory hurdles: Stringent safety regulations and rigorous validation is necessary for widespread adoption by consumers which is not possible by Drive by wire systems.
\end{itemize}

\subsubsection{ Technological Foundation of AI-Enhanced EPS} :
With the evolution of EPS systems, electric motors, actuators and sensors gained traction. This replaced traditional controllers.
For steering purposes , one needs the following data:speed of the vehicle, torque of the vehicle, current position to localize the vehicle and last but not the least, some input either for the driver or in an autonomous vehicle, from software. To achieve this, we design an algorithm which uses ML concepts to fetch the data in real time, analyze it, give input to steering ECU, enable monitoring and detect faults if any. The popular algorithms are based on reinforcement learning and deep learning such as Neural Nets. These algorithms have to satisfy various ODD(Operational Design Domain) Conditions.
\section{Effects on Vehicle Safety through AI in EPS}
\begin{itemize}
\item  Collision Avoidance and Lane Keeping Assistance: 
AI-driven EPS plays a pivotal role in active safety by enabling features like lane-keeping assistance and collision avoidance. Machine learning models can analyze vehicle speed, road markings, and traffic flow to maintain the vehicle within the lane and adjust the steering in real time if an obstacle is detected. Drive by perception is the technology that can be used for this or through pre developed MAPS. This capability significantly reduces the risk of accidents due to human error, especially in complex traffic environments.
\item  Improved Stability and Traction Control :
Real-time processing of sensor data allows AI-enhanced EPS to modulate steering input based on traction conditions. For example, in slippery conditions, the AI system can reduce steering input to prevent oversteer or understeer, enhancing vehicle stability. This is particularly beneficial for emergency maneuvers where quick and precise control is required to maintain safety.
\item Fatigue Detection and Driver Assistance
AI-based steering systems are also evolved in driver monitoring, detecting signs of fatigue or distraction by analyzing subtle changes in steering patterns. Upon identifying unusual patterns based on the data in its memory, the system can issue alerts or adjust control sensitivity, providing an added layer of safety. This can be done through blaring sounds, lights or an automated voice warning the driver.
\end{itemize}
\section{Performance Capabilities of AI-Driven EPS}
\begin{itemize}
\item Adaptive Steering Based on Driving Conditions :
AI gives the capability to the user to adjust steering based on an individual's needs and environmental conditions. One needs to be responsive in heavy traffic conditions specially during evenings in stop and go traffic. On highways , one needs a controlled feeling. We can adapt the steering functionality using AI for any of these conditions ensuring safety and comfort.
\item Predictive Maintenance and System Diagnostics
Many times we see our tire has lost pressure or a tire puncture in the middle of a drive to a hilly area for a weekend getaway. This is frustrating as well as inconvenient. One can predict these situations by continuous monitoring of sensor data. Breakdowns cannot be avoided completely but definitely can be predicted to some extent.
\end{itemize}
\section{Conclusion and Future strategy}
To sum up, the use of machine learning algorithms in EPS is an important progress of automotive technology with AI. The advanced driver assist systems that are part of the modern car will only become smarter as AI technology evolves and advances, leading to much more intelligent magnetic steering-assist functions in future vehicles. Through this paper, we take a look at how the integration of AI into Electric Power Steering is transforming vehicle safety, performance and user experience. AI-supported EPS: The artificial intelligence allows improved driver assistance systems, stability and adaptability by processing predictions in real time. But these advantages are overwhelming provided the cybersecurity, ethical and power requirements still have some challenges to overcome with AI-driven EPS systems. As AI and machine learning evolve, these systems will only get better over time and lay the groundwork for vehicles that are even safer, more efficient, and increasingly autonomous.
Continuous development in edge computing, federated learning, and improved sensor integration will continue to advance AI-driven EPS systems. For instance, edge computing can reduce latency by enabling onboard data processing, enhancing the responsiveness of EPS in real-time scenarios. Federated learning may allow vehicles to share anonymized data, thus allowing improvements in machine learning models collectively across a fleet of vehicles without the compromise of privacy. Future research may also be directed at developing algorithms with a lower computational load to share energy efficiency with performance. Since EPS is becoming the core technology of autonomous vehicles, further studies should be conducted in order to ensure that AI-driven steering systems meet the exacting standards for safety and reliability when deployed en masse.

\end{document}